\documentclass[10pt,conference]{IEEEtran}
\usepackage[numbers]{natbib}
\usepackage{hyperref}
\usepackage{graphicx} 
\usepackage{amsmath} 
\usepackage{balance}

\title{Generational Replacement and Learning for High-Performing and Diverse Populations in Evolvable Robots}
\author{\IEEEauthorblockN{1\textsuperscript{st} K. Ege de Bruin}
\IEEEauthorblockA{\textit{Department of Informatics} \\
\textit{University of Oslo}\\
Oslo, Norway \\
eged@uio.no}
\and
\IEEEauthorblockN{2\textsuperscript{nd} Kyrre Glette}
\IEEEauthorblockA{\textit{RITMO, Department of Informatics} \\
\textit{University of Oslo}\\
Oslo, Norway \\
kyrrehg@ifi.uio.no}
\and
\IEEEauthorblockN{3\textsuperscript{rd} Kai Olav Ellefsen}
\IEEEauthorblockA{\textit{Department of Informatics} \\
\textit{University of Oslo}\\
Oslo, Norway \\
kaiolae@ifi.uio.no}
}

\begin{document}

\maketitle

\begin{abstract}
    Evolutionary Robotics offers the possibility to design robots to solve a specific task automatically by optimizing their morphology and control together. However, this co-optimization of body and control is challenging, because controllers need some time to adapt to the evolving morphology - which may make it difficult for new and promising designs to enter the evolving population. A solution to this is to add intra-life learning, defined as an additional controller optimization loop, to each individual in the evolving population. A related problem is the lack of diversity often seen in evolving populations as evolution narrows the search down to a few promising designs too quickly. This problem can be mitigated by implementing full generational replacement, where offspring robots replace the whole population. This solution for increasing diversity usually comes at the cost of lower performance compared to using elitism. In this work, we show that combining such generational replacement with intra-life learning can increase diversity while retaining performance. We also highlight the importance of performance metrics when studying learning in morphologically evolving robots, showing that evaluating according to function evaluations versus according to generations of evolution can give different conclusions.
\end{abstract}

\section{Introduction}
Traditionally, robots are designed manually by engineers to perform specific tasks. While methods like reinforcement learning are often used to optimize control systems, the morphology of most modern robots is still largely human-designed. This approach can result in efficient robots for predefined tasks, but it limits the robot’s ability to adapt to new tasks or environments, as their morphology is not optimized for adaptability beyond its initial design.

In nature, we see many organisms that can quickly adapt to new environments and solve a wide range of tasks. This is the inspiration for evolutionary robotics, where robots are being designed using evolutionary algorithms \cite{nolfi2016evolutionary}. A somewhat unique problem for robots is the co-optimization of control and morphology \cite{Cheney2016}. The evolutionary algorithm needs to traverse a more complex search space due to the intertwined body and brain. Consequently, there might be a mismatch between control and robot morphology, and a robot's morphology might be discarded because it has not been performing to its full potential \cite{eiben2013triangle}. This causes the algorithm to stagnate quickly, without exploring much of the design space, because new potentially well-performing robot morphologies are being discarded.

\begin{figure}
    \centering
    \includegraphics[width=1\linewidth]{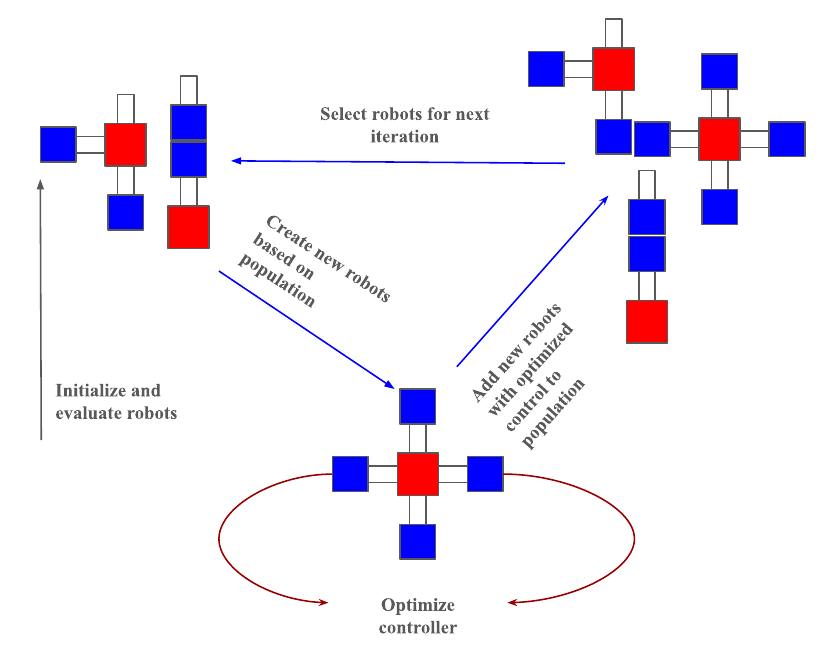}
    \caption{
        Overview of an evolutionary loop for morphology optimization (the blue lines) with a learning loop for controller optimization (the red lines). When comparing over generations, we compare over the blue loop, while comparing over function evaluations also takes the red loop into account.
    }
    \label{figure-evo-learn}
\end{figure}

A common approach to deal with this problem is to make a robot adapt its control to its new morphology, therefore increasing the chance of performing to its morphology's potential. This is often done by adding an inner learning loop to the evolutionary algorithm: An outer evolutionary loop searches through the morphology search space, while an inner learning loop tries to find optimal control parameters for every new robot. An overview of this can be seen in Figure \ref{figure-evo-learn}, where the blue loop is the outer evolutionary loop and the red loop is the inner learning loop. Adding a learning loop has been shown to be an efficient way to find well-performing robots \cite{miras_evolving-controllers_2020, luo_effects_2022, zhao_robogrammar_2020, gupta_embodied_2021, eiben_if_2020}. It can also be improved using Lamarckian principles \cite{luo_enhancing_2023, jelisavcic_analysis_2017, Jelisavcic2019}, where learned behaviours are coded back to the genotype to be inherited by offspring robots.

When comparing evolution with or without learning, it is not obvious how to compare performance fairly. This is because adding learning will require more function evaluations per robot. One can compare over generations, which is proportional to the number of different robot morphologies evaluated, or over function evaluations, which is the number of times a robot is evaluated in the environment. Comparison over generations makes sense because the time it takes to create a robot in real life exceeds the evaluation time of a robot \cite{moreno2022out}. However, in simulation, it makes more sense to compare over function evaluations because in simulation, evaluating a robot takes more time. Compared over generations, adding learning has a clear benefit of evolution only \cite{miras_evolving-controllers_2020}, while over function evaluations, this is not the case \cite{luo_effects_2022}.

Another effect of the mismatch between morphology and control in evolutionary robotics is the lack of diversity \cite{Cheney2016}. New robot morphologies in an evolutionary algorithm that are significantly different from existing robots have a high chance of underperforming because their control is not optimized for it. This makes it difficult to find new robot morphologies. Some possible approaches to tackling this problem have been proposed, such as innovation protection \cite{Cheney2018} or diversity promotion using quality diversity approaches such as MAP-Elites \cite{nordmoen_map-elites_2021}. These approaches need an explicit diversity measure to steer the algorithm towards more diverse solutions, and choosing this measure poses another challenge in designing the algorithm~\cite{Norstein2022}.

Maintaining diversity in evolving robotic populations can also be achieved by selecting an appropriate survivor selection method. Often an elitist approach is used, where the best robots survive \cite{Cheney2016, miras_evolving-controllers_2020, luo_effects_2022, zhao_robogrammar_2020}. However, an approach where offspring robots replace the oldest robots in the population can be used to improve diversity \cite{gupta_embodied_2021}, at the cost of risking losing the best solutions so far. 

In evolutionary algorithms, replacing old solutions with newer ones can be done in a steady-state or full-generation method. Generational algorithms replace the entire population with offspring, such as all offspring replacing parents. Steady-state approaches, in contrast, produce fewer offspring per cycle, allowing incremental replacements. While steady-state methods, such as the replace-oldest strategy employed by Gupta et al. \cite{gupta_embodied_2021}, are effective in dynamic environments due to faster adaptation and early convergence \cite{vavak1996comparative}, they risk falling into local optima \cite{hancock1994empirical}. Generational replacement maintains diversity, often outperforming steady-state in static environments. 

Generational replacement emphasizes the mismatch between control and morphology. In this paper, we hypothesize that generational replacement will perform better when combined with intra-life controller learning. Learning and a survivor selection that removes the oldest individuals from the population have also been combined in previous work \cite{gupta_embodied_2021, faina2013edhmor}, but to our knowledge, no previous work has systematically analyzed these factors (evolution with and without learning, selection with and without elitism) to uncover their full relationship and effect on morphologically evolving robots.

Our main contributions are:
\begin{itemize}
  \item A demonstration that the combination of learning and a generational survivor selection leads to populations that are both high-performing and diverse.
  \item Highlighting the importance of evaluation metrics when comparing morphologically evolving robots with and without learning: When measured over \emph{generations}, we see a clear benefit of learning, but this benefit is much less clear when we instead measure over \emph{function evaluations}.
  \item Initial evidence that the role of learning is more complex than previous studies have indicated: We find that adding a learning phase \textit{after} evolution can give benefits close to those of interleaving evolution and learning \cite{etor2024}.
\end{itemize}
The code and videos are available on Github\footnote{See \url{https://tinyurl.com/ssci2025-egedebruin}}{}.

\section{Setup}

\begin{figure}
    \centering
    \includegraphics[width=0.7\linewidth]{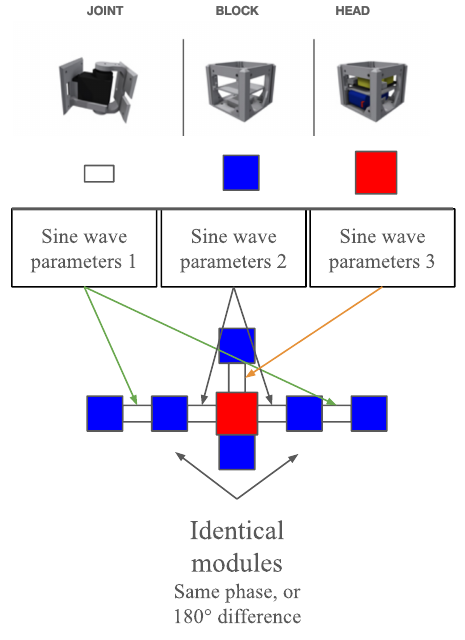}
    \caption{
        Overview of the robot's control and morphology. There are three types of modules, and every robot always has one head module. The head module has four attachment points, the block module has six (it includes up and down) and the joint module has two. Symmetry is ensured by making the left and right parts of the core module identical. Each joint module is controlled by its own sine wave parameters, with symmetrical parts sharing the same set of parameters.
    }
    \label{figure-overview}
\end{figure}

\subsection{Robot phenotype}
For this work, we make use of Revolve2\footnote{See \url{https://github.com/ci-group/revolve2}}{} as a modular robot framework, where the modules are based on the RoboGen framework \cite{Robogen}. The MuJoCo rigid body physics simulator is used to simulate the robots. For future work, it is possible to 3D print these modules, attach them and evaluate the robots in real life. The morphologies are built by attaching building blocks using three different types of modules. These are a head module, a block module, and a joint module as seen in Figure \ref{figure-overview}. Every robot has exactly one head module; the evolutionary algorithm decides the amount and position of the other two types of modules. The head module has four attachment points, the block has six, the joint has two, and the joint module is the only movable module. These joint modules are actuated by a sine-wave controller for each hinge separately. It is, therefore, a decentralized controller approach. The following equations are used for control:
\begin{equation}
\begin{split}
\Theta_i & = A * sin(\phi_i + P) + O \\ \phi_i & = \phi_{i-1} + \Delta_\phi * F
\end{split}
\end{equation}

In these equations, $\Theta_i$ is the output of the controller at time step $i$, $A$ is the amplitude, $P$ is the phase offset and $F$ is the frequency. $A$ and $P$ are the two learnable parameters, and $F$ is set to 4. Even though the control is decentralized, hinges might share parameters due to symmetry, which will be explained in the next subsection.

\begin{figure}
    \centering
    \includegraphics[width=1\linewidth]{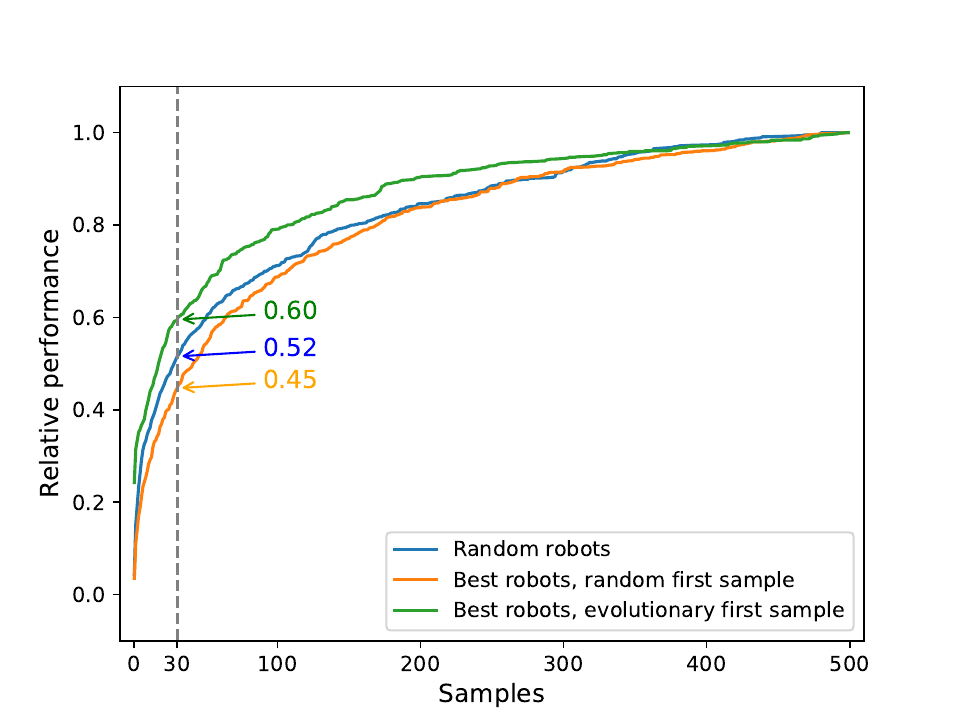}
    \caption{
        To find a suitable learning budget for Bayesian optimization, we tested how fast robot performance increases during the search. We tested Bayesian optimization on 100 instances of two types of robots and plotted their performance relative to their maximum performance. The two types of robots are random robots and the best robots of a preliminary evolutionary search. We plot two lines for the best robots, one with a random first sample and one with the first sample of the evolutionary search. After 30 samples, on average already 50\% of the robots' potential is reached, this learning budget is therefore chosen for the evolution-with-learning setting.
    }
    \label{figure-bo}
\end{figure}

\subsection{Robot genotype}
The robot morphology is directly encoded in the genotype and can therefore be represented as a tree. Every robot has a head module and the first robots are initialized by iteratively adding random modules in random positions. We ensure symmetry by making the left and right sides of the core module identical. That makes that there are three attachment points to the head module in the genotype, one for the front, one for the back, and one for the left and right sides of the robot. Every joint module in the genotype has a sine wave as a controller, and this also means that a joint on the left side of the core module has the same sine wave parameters as its equivalent on the right side due to symmetry. We enforce symmetry to resemble real-life creatures more. Ideally, the symmetrical parts move identically, and therefore, they share parameters. 

There are three possible mutations to the genotype, of which the first two are influenced by earlier work using direct encodings \cite{gupta_embodied_2021}: Firstly, a random number of modules can be added to the robot. Secondly, a random number of modules can be removed from the robot. Adding and removing a module can be done at any place in the genotype, so not necessarily only the leaves. This can change the morphology more drastically but can help the search for new places in the search space. Lastly, there is a parameter that decides whether the symmetrical parts of the robot are in the same phase or an alternating phase, and the third mutation switches this. We add the possibility of an alternating phase between the symmetrical parts to make a walking-like pattern possible. As mentioned before, the parameters for the controller of an actuated hinge are the amplitude and phase offset of the sine wave, and a mutation changes these parameters by adding or subtracting Gaussian noise from these parameters. When a new joint is created, a new controller is made with random parameters, and when one is removed, this controller is also removed. An overview of the robot genotype and phenotype can be seen in Figure \ref{figure-overview}.

\subsection{Learning method}

\begin{figure}
    \centering
    \includegraphics[width=0.5\linewidth]{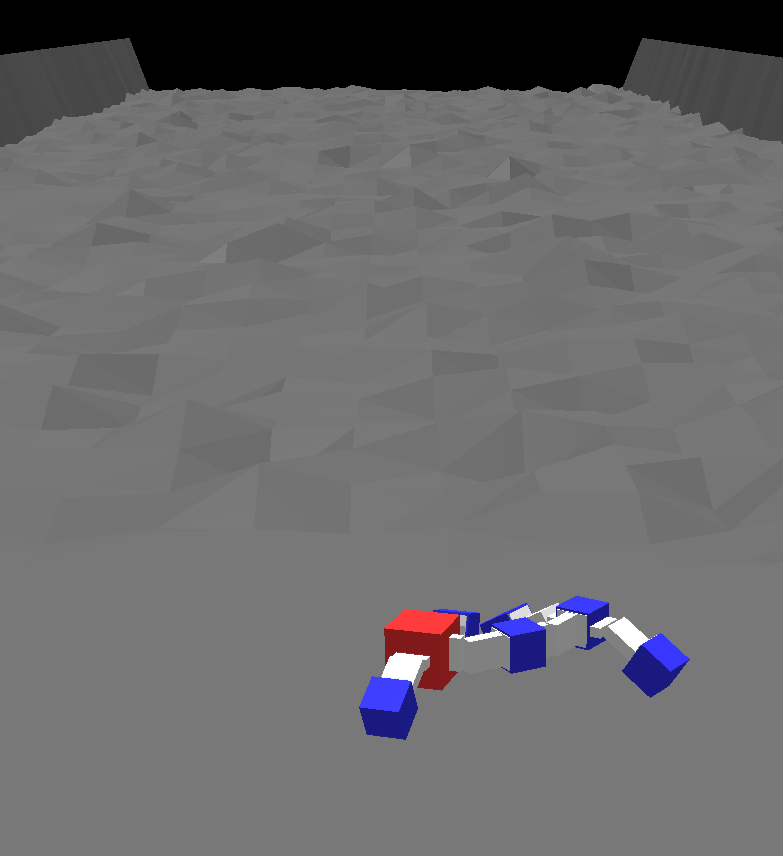}
    \caption{
        The rough environment with an example robot. The environment is generated by adding noise to a flat height map, and the height map is the same for every evaluation. The goal is to move forward as far as possible in 30 simulated seconds.
    }
    \label{figure-environment}
\end{figure}

The algorithm used for learning is Bayesian optimization, chosen for its fast learning capability and sample efficiency \cite{lan_learning_2021, vanDiggelen2021, legoff2023}. While Bayesian optimization has high computational complexity, this is negligible in our experiments due to the small number of controller parameters and the limited learning budget. Each robot's Bayesian optimization process starts with the control parameters obtained from the evolutionary algorithm, potentially accelerating the learning process.

We employ the Matern 5/2 kernel with a length scale of 0.2 and use the Upper Confidence Bound as the acquisition function with an exploration parameter of 3, a configuration previously shown to be effective in evolving robots for directed locomotion \cite{lan_learning_2021, vanDiggelen2021}. Two scenarios are compared: one without learning and one with a learning budget of 30. In the no-learning scenario, Bayesian optimization is omitted, and the robot morphology's performance is assessed using the evolved control parameters from a single trial.

To determine the appropriate learning budget, we sampled 100 random robots and 100 robots optimized through evolutionary search, running Bayesian optimization on each until no further improvement was observed. As shown in Figure \ref{figure-bo}, the algorithm achieved approximately 50\% of its potential after 30 samples on average. When another controller type and learning algorithm is used, this can then be used to decide the learning budget.

\begin{figure}
    \centering
    \includegraphics[width=1\linewidth]{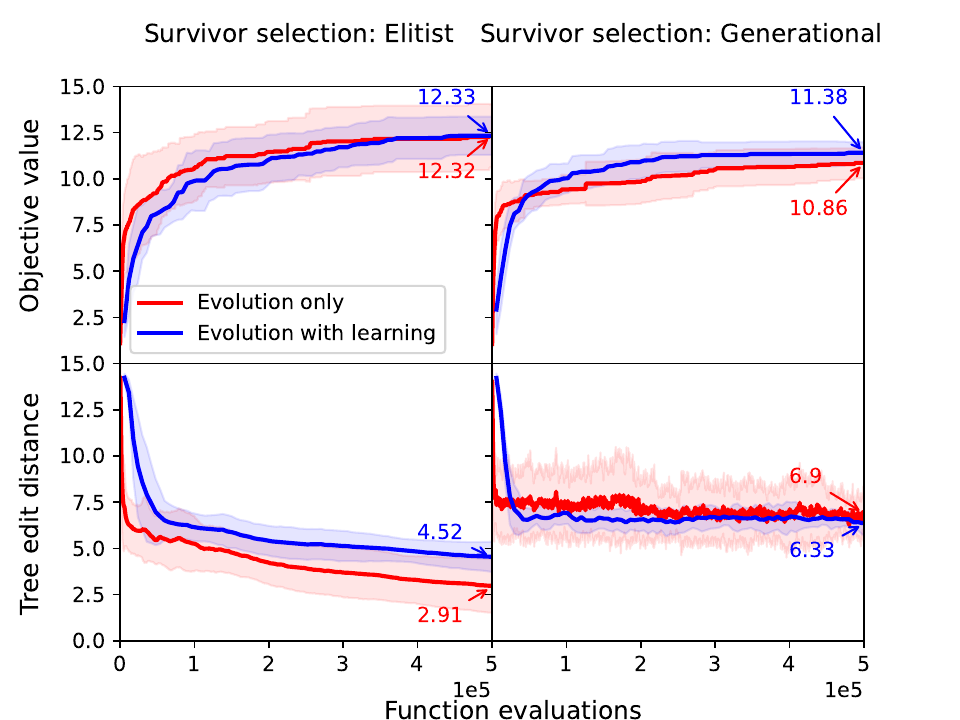}
    \caption{
        The performance (objective value) and diversity (tree edit distance) compared over function evaluations. The performance shows the best objective value so far. The values at the end are indicated for clarity. The lines are averaged over runs, and the shaded areas are the standard deviation. The initial jump in performance and drop in diversity can be explained by the algorithm initially filtering out bad morphologies.
    }
    \label{figure-line-fe}
\end{figure}

\subsection{Evolutionary algorithm}

The evolutionary algorithm explores the search space of robot morphologies to identify the best designs for the task. The population size is 200, and each new robot undergoes a learning phase to optimize its control parameters. In each generation, 200 offspring robots are created. Offspring are generated via asexual reproduction, with parent selection using tournament selection and a tournament size of 20. We use a high selection pressure to keep a high proportion of offspring of well-performing robots in the population with generational replacement.

Robots' sizes in the initial population range from 15 to 20 modules. Mutations can add or remove up to three modules, but the maximum size is capped at 20 modules. If a mutation exceeds this size, a new mutation is attempted. This size limit ensures the robots remain feasible for physical construction. Preliminary experiments showed that without a cap, robots after an add-mutation would be selected more often and cause robots to grow continuously during the evolutionary search. We suspect this is because adding a module potentially changes the least in the robot morphology. While growing morphologies in evolutionary robotics is an interesting area of research \cite{kriegman2018morphological}, it is beyond the scope of this work.

We use a rough environment, as illustrated in Figure \ref{figure-environment}, which remains consistent across evaluations. The objective is to maximize forward displacement over 30 seconds. The following evolutionary settings are evaluated: evolution-only vs. evolution-with-learning and elitist vs. generational survivor selection, resulting in a total of four parameter configurations. In elitist selection, the top 200 robots from the combined parent-offspring pool survive, while in generational selection, the offspring survive. Each configuration is tested across 12 repeated trials. We perform Mann-Whitney U-tests to test significant differences between configurations.

\section{Results}

\subsection{Elitist and Generational}

\begin{figure}
    \centering
    \includegraphics[width=1\linewidth]{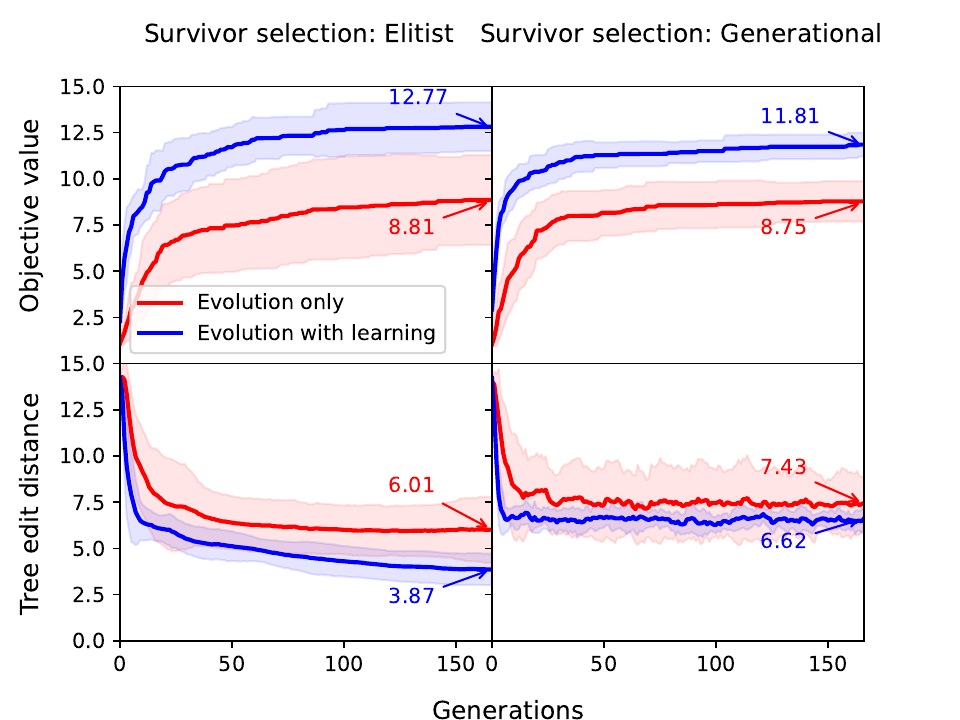}
    \caption{
        The performance (objective value) and diversity (tree edit distance) compared over generations. The performance shows the best objective value so far. The values at the end are indicated for clarity. The lines are averaged over runs, and the shaded areas are the standard deviation. We end the plot at 166 generations, as that is as far as evolution with learning goes. There are more generations for the evolution-only case.
    }
    \label{figure-line-gen}
\end{figure}

We first compare the approaches based on function evaluations, with results shown in Figure \ref{figure-line-fe}. The performance plots track the best-so-far performing robot rather than the best-performing robot in the current population. The reasoning is that the main goal is to identify a high-performing robot; the stage in the algorithm where this robot is found is less important. We use the average tree edit distance within the population as diversity measure \cite{samuelsen2014some}.

The results show that the evolution-only approach with elitist survivor selection leads to poor diversity, while using generational survivor selection reduces performance. However, incorporating learning improves results in both cases. With elitist survivor selection, evolution with learning increases diversity while maintaining the same level of performance as the evolution-only approach. With generational survivor selection, diversity remains high ($p < 0.001$ for evolution only and evolution with learning), and the performance of generational replacement with learning is better than the evolution-only approach, even though the difference is not significant ($p > 0.05$). As previously mentioned, these plots show the best-so-far solution, so the final population in generational replacement with learning has better-performing robots than without learning.

Generational replacement combined with learning provides the best balance: low selection pressure fosters high population diversity while learning ensures that these diverse robots achieve good performance. However, learning does not improve performance as much as expected compared to the evolution-only approach ($p > 0.05$ for elitist and generational replacement). 

\subsection{Comparing over generations}
Figure \ref{figure-line-gen} shows the results compared over generations. Evolution with learning has a significantly better performance than without learning ($p < 0.001$ for both elitist and generational), and this is similar to what has been seen before in previous works \cite{miras_evolving-controllers_2020, luo_effects_2022}. This difference also makes sense, because evolution with learning has had 30 times as many function evaluations to test controller parameters. The elitist survivor selection has a better performance than generational replacement after 166 generations ($p < 0.05$ for evolution with learning), although the difference for evolution-only is not significant ($p > 0.5$).

Compared over function evaluations it was shown that the diversity for the elitist approach with learning is higher than without learning ($p < 0.05$), here the diversity plot shows the opposite ($p < 0.005$). The generational survivor selection approach still has higher diversity ($p < 0.001$ for evolution with learning), even though the difference is not significant for evolution-only ($p > 0.05$). 

\begin{table}
\caption[Caption]{Performance improvement after an additional learning phase.}
\centering
\begin{tabular}{|l l | c | c |} 
 \hline
  && \begin{tabular}{@{}c@{}}166th \\ generation\end{tabular} & \begin{tabular}{@{}c@{}}Learning \\ phase\end{tabular} \\ 
  \hline
  \begin{tabular}{@{}c@{}}Evolution \\ Only\end{tabular} & Elitist & 8.81 & 11.10 (+2.29) \\
  &Generational& 8.75 & 9.67 (+0.92) \\
  \hline
  \begin{tabular}{@{}c@{}}With \\ Learning\end{tabular} & Elitist & 12.77 & X \\
  &Generational& 11.81 & X \\
 \hline
\end{tabular}
\label{table:1}
\end{table}

To investigate whether the evolution-only morphologies after 166 generations performed poorly due to a mismatch between control and morphology, we added a learning phase to morphologies. We performed an additional Bayesian optimization process for all 200 morphologies in the 166th generation of all 12 runs. The additional learning budget was 500 samples, and the results can be seen in Table \ref{table:1}. For elitism, there is a large increase in performance after the additional learning phase. The additional learning phase reduces the difference in objective value from 3.96 to 1.67. There is an increase in the generational approach as well, but the difference is smaller.

\begin{figure}
    \centering
    \includegraphics[width=1\linewidth]{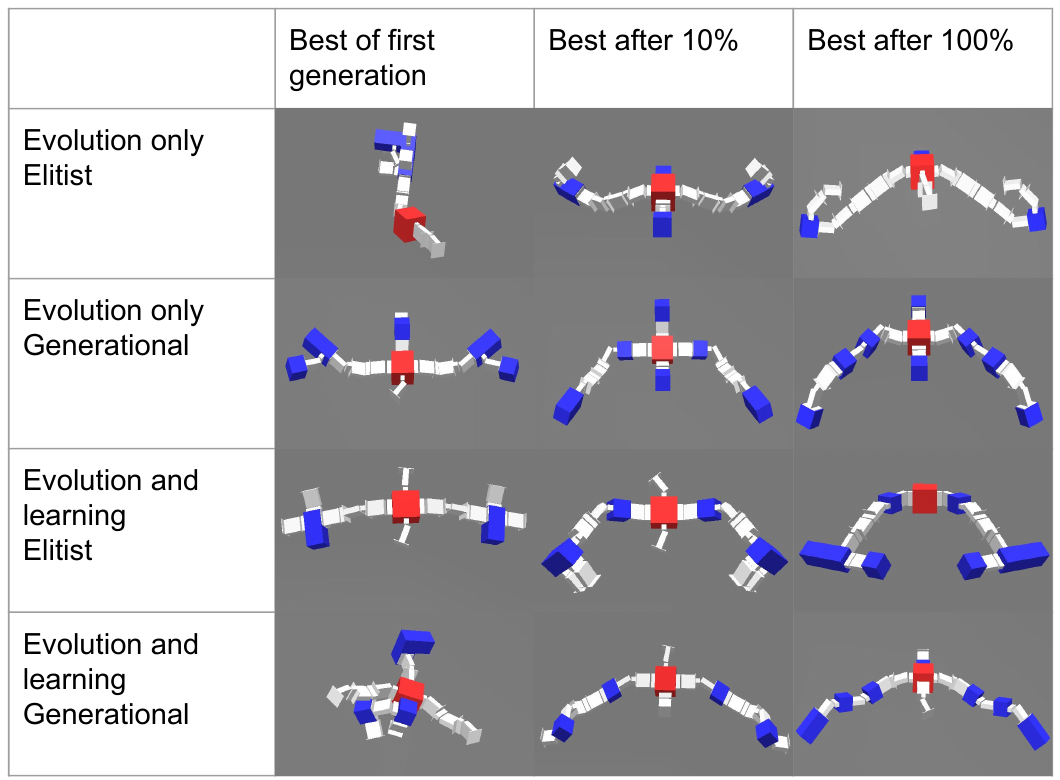}
    \caption{
        Robots from the best run for each of the parameter settings. We show the best of the first generation, the best after 10\% of the evolutionary search and the best after 100\% of the evolutionary search.
    }
    \label{figure-robots}
\end{figure}

In Figure \ref{figure-robots} we show three robots of the best run for each of the parameter settings, each in a different stage of the evolutionary algorithm. After 10\% of the evolutionary search, we see that the robot morphology does not change much anymore compared to after 100\% of the search, but we can still notice some changes. Lastly, as previously mentioned, one of the parameters in the evolutionary search decides whether the symmetrical joints are in the same phase or alternating phases, and the best robots in all runs have the same phase in the symmetrical joints.

\section{Discussion}
While previous work has shown that learning benefits the evolutionary search when evolving robots, our results show a more complex story. Firstly, there is a big difference between comparing over generations and function evaluations. While previous work has shown that learning has great benefits when compared over generations, we argue that in a simulated setting comparing over function evaluations makes more sense, and in this case, there is far less benefit of learning. Our results also show that evolution-only has not converged yet after 166 generations. After 500.000 function evaluations (or 2500 generations) it catches up to evolution with learning. In a real-life setting, where comparing over generations makes more sense, it might not be viable to evolve for 2500 generations, which is why we added a learning phase to the 166th generation of the evolution-only morphologies. In this case, the performance of evolution-only almost catches up with evolution with learning.

When evolving robots, often an elitist strategy is used. We compare it to a generational replacement strategy and the results show that it increases population diversity. The results also show that in this case, adding learning increases performance. The additional learning phase to evolution-only in this case does not catch up with evolution with learning. We showed that a generational survivor selection with learning has a positive impact on both performance and diversity in a population of evolving robots. However, the performance does not equal the setting with elitism yet. A disadvantage of elitism is a reduced diversity in the population, but if the main goal of the search is to find good-performing robots and diversity is treated as a means to an end, the benefits of the generational selection need to be found in performance only. We argue that this can be found in the lower standard deviation seen in Figures \ref{figure-line-fe} and \ref{figure-line-gen}, resulting in a more stable evolutionary search.

We used a direct encoding for the morphology, where every mutation leads to a new morphology. Moreover, there was only asexual reproduction, so there existed no crossover. These decisions impacted the evolutionary search by influencing how different offspring are to their parent solutions. A more neutral mutation strategy will make offspring more equal to their parents and favour an evolutionary search, while crossover could have the opposite effect. Our choice of not having crossover and not having a neutral mutation is intended as a fair middle point on the spectrum of mutability. An interesting venue for future work is to look more closely at the effect of mutability on the phenomena we study, by changing our direct encoding or even replacing it with an indirect encoding.

It is additionally important to note that the learning budget used in this work is quite low. We showed our reasoning for choosing our learning budget, which can be copied for other more complex tasks or robot controllers. However, the simplicity of the environment, task and controller method influences the evolutionary search as well, and the effect of learning might be more prevalent in a setting where an evolutionary search is not enough. This will then require more complex environments and tasks. For the environment used in this work a sine wave is enough and more complex controllers are not needed.
\balance 
\section{Conclusion}
We have investigated the influence of learning on morphologically evolving robots. Previous work on the comparison between evolution-only and evolution with learning focused on an elitist survivor selection, and we have shown the effect of a generational survivor selection. This performs almost as well as elitism while preserving diversity in the populations. Another contribution of this work is that we have shown that the type of comparison is important. Comparing over generations shows a clear benefit of learning, and this comparison makes sense when the cost of creating a robot is higher than evaluating a robot. However, in simulation, evaluating a robot is more resource-intensive than creating one, making it more logical to compare robots based on their function evaluations. We demonstrated that in such cases, the advantage of learning is smaller compared to when the evaluation is done across generations, and that a generational survivor selection is needed to keep diversity. Finally, we added a learning phase to the final generation of the evolution-only setting, and showed that it catches up closely to evolution with learning. This suggests that the robot morphologies have not optimised their control yet and that a learning loop during evolution might not be needed if a learning phase is added after the evolutionary search.

Promising topics for future work include investigating the influence of learning and generational replacement with a different morphological encoding. It would also be interesting to look into more complex tasks, environments or controllers where an evolutionary search would not be enough, showing more benefit of learning. Lastly, it has been shown that Lamarckian inheritance between parent and offspring solution has a positive impact on performance in an elitist selection strategy - investigating if this effect is also present with a generational selection strategy like the one used herein would be an interesting direction for future studies on Lamarckian evolution.

\section*{Acknowledgements}
This work was partially supported by the Research Council of Norway through its Centres of Excellence scheme, project number 262762.

\bibliographystyle{ieeetr}
\bibliography{references}

\end{document}